# Color naming reflects both perceptual structure and communicative need

Noga Zaslavsky[1,4] (noga.zaslavsky@mail.huji.ac.il)
Charles Kemp[2] (ckemp@cmu.edu)
Naftali Tishby[1,3] (tishby@cs.huji.ac.il)
Terry Regier[4,5] (terry.regier@berkeley.edu)

[1]Edmond and Lily Safra Centre for Brain Sciences, The Hebrew University, Jerusalem 9190401, Israel
[2]Department of Psychology, Carnegie Mellon University, Pittsburgh, PA 15213 USA
[3]The Benin School of Computer Science and Engineering, The Hebrew University, Jerusalem 9190401, Israel
[4]Department of Linguistics and [5]Cognitive Science Program, University of California, Berkeley, CA 94720 USA

**Abstract**

Gibson et al. (2017) argued that color naming is shaped by patterns of communicative need. In support of this claim, they showed that color naming systems across languages support more precise communication about warm colors than cool colors, and that the objects we talk about tend to be warm-colored rather than cool-colored. Here, we present new analyses that alter this picture. We show that greater communicative precision for warm than for cool colors, and greater communicative need, may both be explained by perceptual structure. However, using an information-theoretic analysis, we also show that color naming across languages bears signs of communicative need beyond what would be predicted by perceptual structure alone. We conclude that color naming is shaped both by perceptual structure, as has traditionally been argued, and by patterns of communicative need, as argued by Gibson et al. – although for reasons other than those they advanced.

## Introduction

Languages vary widely in the ways they partition colors into categories. At the same time, this variation is constrained, and similar color naming systems are often seen in unrelated languages (e.g. Berlin & Kay, 1969; Lindsey & Brown, 2006). The forces that give rise to this constrained variation have long been debated, and it is often held that a major role is played by perceptual structure (e.g. Kay & McDaniel, 1978). A variant of this view emphasizes in addition the importance of communicative forces, and argues that languages divide perceptual color space into categories in ways that support efficient communication (Jameson & D'Andrade, 1997; Regier et al., 2007; Baddeley & Attewell, 2009; Regier et al., 2015; Lindsey et al., 2015).

Recently, Gibson et al. (2017) suggested an even greater role for communicative forces. They proposed that cross-language commonalities in color naming may reflect a human need to refer to particular colors more than others, and presented this hypothesis as an alternative to one based on perceptual salience (p. 10785). They showed that color naming systems across languages support more precise communication about warm colors than cool colors, and that the objects we talk about tend to be warm-colored rather than cool-colored — suggesting that color naming systems may have adapted to a general human need to communicate preferentially about warm colors.

Here, we engage this argument, and present results that suggest a somewhat different conclusion. We first present the core of Gibson et al.'s argument in detail, and replicate their findings. We then consider an alternative explanation of their findings, and show that greater communicative precision for warm than for cool colors, and greater need for warm colors, may both be explained by perceptual structure, without any additional communicative preference for warm colors. We next present a novel information-theoretic analysis of the link between need and communicative precision, and we use that analysis to infer need from color naming data. On that basis, we show that color naming across languages bears signs of communicative need beyond what would be predicted by perceptual structure alone. We conclude that color naming is shaped both by perceptual structure, as has traditionally been argued, and by patterns of communicative need, as argued by Gibson et al. — although our reasons for implicating need are different from theirs.

## The argument of Gibson et al. (2017)

Gibson et al. found that across languages warm colors tend to be communicated more precisely than cool colors. They also found that the objects we talk about tend to be warm-colored rather than cool-colored, and in that sense warm colors have higher communicative need. They concluded that the warm-cool asymmetry in communicative precision across languages "reflects colors of universal usefulness" and that the principle of color use "governs how color categories come about" (p. 10785). They presented this idea as an alternative to proposals based on perceptual salience (p. 10785). Below we present the data they considered, and their definitions of communicative precision and communicative need, which inform our own analyses.

**Data.** Gibson et al. based their analysis primarily on color naming data from the World Color Survey (WCS: Cook, Kay, & Regier, 2005). The WCS dataset contains color naming data from 110 languages of non-industrialized societies. In the WCS, native speakers of each language were asked to provide a name for each of 330 color chips. Gibson et al. analyzed naming data for the subset of 80 color chips shown in Figure 1, for all WCS languages and also for three languages for which they collected data: English, Spanish, and

Tsimané. For each language $l$, each color term $w$ in $l$, and each color chip $c$, they estimated the color naming distribution $p_l(w|c)$ as the proportion of speakers of $l$ who used $w$ rather than some other term to name $c$.

**Communicative need.** A need distribution, reflecting how often a given color $c$ is used in communication (Regier et al., 2015), can be naturally considered a prior distribution $p(c)$ over colors. Gibson et al. considered two priors: a uniform prior, and a "salience-weighted prior" (p. 27 of SI). In the salience-weighted prior, the probability of each color was determined by the proportion of times that color appeared in a foreground object, rather than in the background, in their study of natural images. This prior was based on the assumption that foreground objects are more likely to be talked about than are backgrounds. This salience-weighted prior exhibits greater probability mass for warm colors than for cool colors (see Figure 4C).

**Communicative precision.** Gibson et al. considered the expected surprisal of a given color $c$, with respect to a color naming distribution $p(w|c)$ and a prior $p(c)$, defined by

$$S(c) = -\sum_w p(w|c) \log p(c|w), \quad (1)$$

where $p(c|w)$ is obtained by applying Bayes' rule:

$$p(c|w) = \frac{p(w|c)p(c)}{\sum_{c'} p(w|c')p(c')}. \quad (2)$$

Lower values of $S(c)$ correspond to higher communicative precision for a given color $c$. Gibson et al. found that across languages $S(c)$ tends to be lower for warm colors (reds/yellows) than for cool colors (blues/greens), when evaluated either with the uniform prior or with the salience-weighted prior. We replicated these results on very similar data (the WCS+ dataset; see below) for both priors, as shown in Figure 3A and Figure 3B.

Notice that $S(c)$ depends both on the prior $p(c)$ and on the naming system $p(w|c)$, and thus these results are an outcome of the combination of need and language. Here we further explore the nature of this combination in two ways: first by using the same priors as Gibson et al. while considering new hypothetical color naming data, and second by keeping the color naming data fixed and considering new priors.

## The role of perceptual structure

The crux of Gibson et al.'s argument is that the warm-cool asymmetry in precision may reflect the warm-cool asymmetry in need. Another possibility, however, is that both asymmetries may be produced by a common underlying cause, perhaps perceptual in nature. Figure 2 re-plots the 80 colors from Figure 1 in CIELAB color space, in which the Euclidean distance between nearby colors corresponds roughly to their perceptual dissimilarity (Brainard, 2003; but see also Komarova & Jameson, 2013). This visualization shows that there exist potentially relevant perceptual asymmetries of color — and in fact this perceptual structure has been used to explain patterns of color naming across languages (Jameson & D'Andrade, 1997; Regier et al., 2007, 2015; Zaslavsky et al., 2018). We wished to understand whether the structure of perceptual color space could also explain the asymmetry in precision documented by Gibson et al., or that in need, or both – a possibility acknowledged by Gibson et al. (p. 10789).

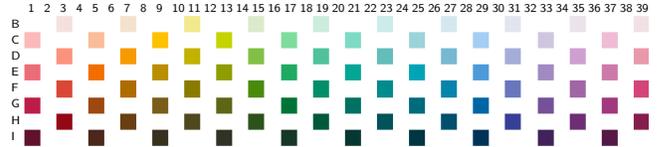

Figure 1: The 80 color chips analyzed by Gibson et al. (2017), represented in the standard WCS palette. White spaces indicate WCS chips that were excluded from the analysis. The achromatic WCS color chips were also excluded.

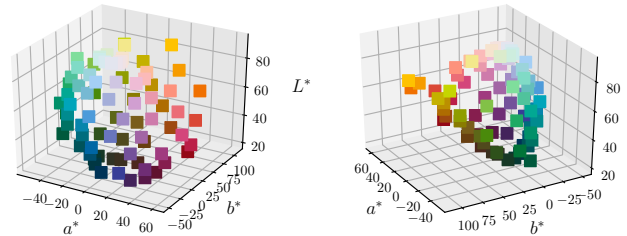

Figure 2: The 80 color chips of Figure 1, represented in CIELAB color space. $L^*$ corresponds to lightness, and hue and saturation are represented in polar coordinates in the orthogonal plane defined by $a^*$ and $b^*$. The irregular distribution of these colors reflects a perceptual asymmetry between warm and cool colors.

To test whether perceptual structure accounts for the warm-cool precision asymmetry, we considered a set of hypothetical color naming systems that were derived solely from the structure of color space, without any additional element of communicative need. We began with the color naming data of the WCS, supplemented by data for English (Lindsey & Brown, 2014); we call this joint dataset WCS+. We considered the same 80 chips used by Gibson et al. Then for each actual language $l$, we constructed a corresponding hypothetical system by clustering the 80 color chips into $k_l$ categories, using the k-means algorithm with respect to the Euclidean distance between colors in CIELAB space. We took $k_l$ to be the number of color terms in language $l$ for which at least two speakers used that term to name the same color chip. In an attempt to avoid local optima, we ran the k-means algorithm 30 times for each language and retained the best solution. This procedure yielded a set of artificial color naming systems that are comparable in number of terms to those in our cross-language data but are determined only by the structure of perceptual color space, with no additional element of need.

The lower panels of Figure 3 show that these k-means systems exhibit a warm-cool surprisal asymmetry broadly simi-lar to that in the actual languages, both with the uniform prior (Figure 3C) and with the salience-weighted prior (Figure 3D). In support of this qualitative observation, with the salience-weighted prior, we found a strong correlation ($r = 0.73$, $p < 0.0001$) between $S(c)$ averaged across actual languages and $S(c)$ averaged across the corresponding k-means systems. With the uniform prior, although an overall warm-cool asymmetry is visually apparent, there is also a clear discrepancy between the actual languages and the k-means systems: light colors tend to have relatively low surprisal in the actual languages, but high surprisal in the k-means systems. In this case we did not find a significant correlation between average surprisal across actual and k-means systems when considering all color chips, but we did find a significant correlation ($r = 0.57$, $p < 0.0001$) when focusing specifically on warm and cool colors by excluding the chips in rows 'B' and 'I' in Figure 1A, which correspond roughly to light and dark. These results suggest that the warm-cool precision asymmetry found for actual languages under Gibson et al.'s priors may to some extent reflect perceptual structure.

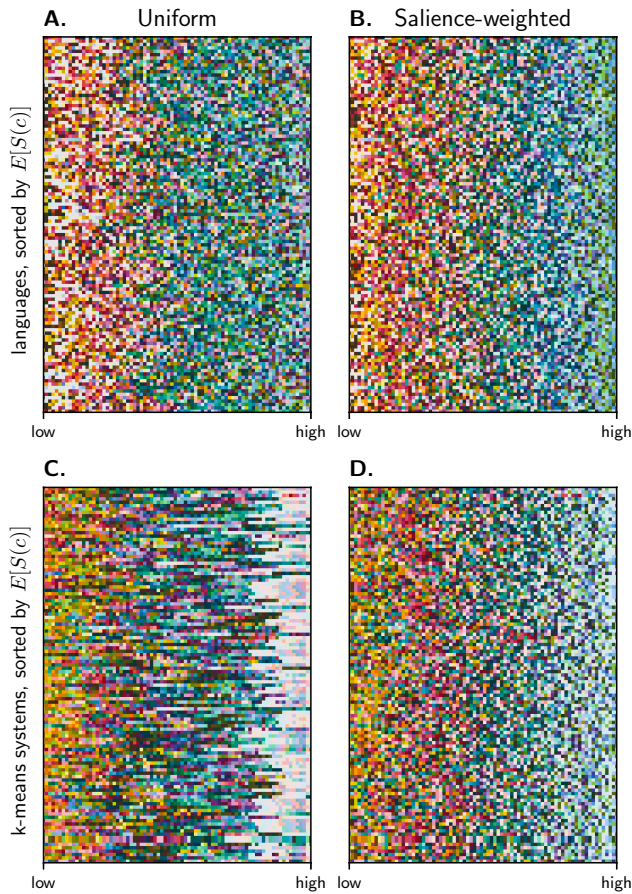

Figure 3: **A-B.** Replication of the results reported by Gibson et al. (2017) for the uniform prior and salience-weighted prior. Across languages, warm colors have lower expected surprisal than cool colors. **C-D.** Analogous analyses in which each language's color naming system was replaced by a hypothetical color naming system obtained by k-means clustering of the color chips represented in CIELAB space. These perceptually-derived hypothetical systems also exhibit a warm-cool surprisal asymmetry.

Perceptual structure may also explain the pattern of color use or need that Gibson et al. reported and captured in their salience-weighted prior itself, according to which foreground objects (as opposed to their backgrounds) are more likely to be warm-colored rather than cool-colored. We found that their salience-weighted prior is correlated ($r = 0.49$, $p < 0.0001$) with the distance of each chip from central gray in CIELAB space,[1] suggesting that the salience-weighted prior reflects how "un-gray" and thus perceptually salient different colors are. It is possible that useful objects are often saliently (warmly) colored so as to attract human attention.

Taken together, these results suggest a possible perceptual common cause for both of the qualitative asymmetries in communicative precision and communicative need that Gibson et al. documented. However these results still leave open the possibility that color naming across languages may be shaped by an element of need beyond what is predicted by perceptual structure. In the following sections we demonstrate an information-theoretic link between communicative need and precision, and use it to address this open question.

## Information-theoretic link between need and precision

When viewing language in information-theoretic terms, one often considers a communication channel between a speaker and a listener (e.g. Plotkin & Nowak, 2000; Baddeley & Attewell, 2009; Gibson et al., 2013). However, this is not the only potentially relevant channel. From an information-theoretic perspective, any conditional distribution can be interpreted as a channel (Cover & Thomas, 2006), and in the present treatment, the lexicon is captured by the conditional distribution $p(w|c)$, which specifies the probability of using a color term $w$ for a given color $c$. Therefore the lexicon itself can be seen as a channel, and one may explore the capacity of that channel — that is, the maximal amount of information about color that can be conveyed by that lexicon.

Formally, the input to this channel is a color $c$, taken from a set $\mathcal{C}$ of colors, and the output is a word $w$, taken from a set $\mathcal{W}$ of possible words. Here we define $\mathcal{C}$ to be the 80 color chips shown in Figure 1, and $\mathcal{W}$ to be an arbitrary set of $K$ words, where $K$ is determined by the number of color terms in the language. Shannon's channel coding theorem (Shannon, 1948) states that the maximal number of bits on average that can be transmitted per channel use is determined by the channel capacity, which is defined as the maximal mutual infor-

---
[1] We took central gray to be located at the midpoint between the CIELAB coordinates for the two achromatic chips that are most intermediate between black and white in the WCS palette, namely E0 and F0 (not shown in Figure 1).

mation between the input and output, namely by

$$\max_{p(c)} I(W;C), \qquad (3)$$

where the maximization is over all possible choices of $p(c)$, and the mutual information is

$$I(W;C) = \sum_{c,w} p(c)p(w|c) \log \frac{p(c|w)}{p(c)}. \qquad (4)$$

A distribution $p(c)$ over $\mathcal{C}$ that attains the channel capacity, i.e. a maximizer of equation (3), is called a capacity-achieving prior (CAP). In our case, since $\mathcal{C}$ and $\mathcal{W}$ are finite sets, a capacity-achieving prior can be found via the Blahut-Arimoto algorithm (Blahut, 1972; Arimoto, 1972). This algorithm is based on the fact that by differentiating equation (4) with respect to $p(c)$ we get the following necessary and sufficient[2] condition for optimality:

$$p(c) \propto \exp(-S(c)). \qquad (5)$$

We find it interesting that while Blahut and Arimoto derived the expression for $S(c)$ from the capacity achieving principle, the same expression has been used for different reasons by Gibson et al. and others (e.g. Piantadosi et al., 2011). Note that equation (5) defines a self-consistent condition for optimality, because $S(c)$ also depends on the prior. By taking the log on both sides of equation (5) we get that a prior is a CAP if and only if it satisfies

$$-\log p(c) = S(c) + \log Z, \qquad (6)$$

where $Z$ is the normalization factor of equation (5).

Thus, need and communicative precision are linked through the capacity achieving principle. Specifically, for a capacity-achieving prior, i.e. a prior $p(c)$ that maximizes the information about color that is conveyed by a given lexicon, we should see a simple linear relationship, with slope 1, between $-\log p(c)$ and the expected surprisal (or communicative imprecision) $S(c)$. Notice that the link between $p(c)$ and $S(c)$ in equation (6) implies that, ideally, patterns in $p(c)$ would be mirrored in $S(c)$, and thus the link is consistent with Gibson et al.'s findings. However this link makes a stronger claim in that it specifies more precisely what the relation between need and precision should be, and does so on theoretically motivated grounds. In the next section we use this information-theoretic link to present new evidence that color naming across languages may indeed reflect universal patterns of communicative need, as well as perceptual structure.

## Inferring need from naming data

The capacity achieving principle provides a basis for inferring a theoretically-motivated need distribution from color naming data. Concretely, given a color naming system, this principle allows us to infer what the accompanying need distribution or prior should be in order to maximize the precision of the given lexicon.

[2]This follows from the concavity of $I(W;C)$ in $p(c)$. For more detail see Theorem 2.7.4 and section 10.8 in (Cover & Thomas, 2006).

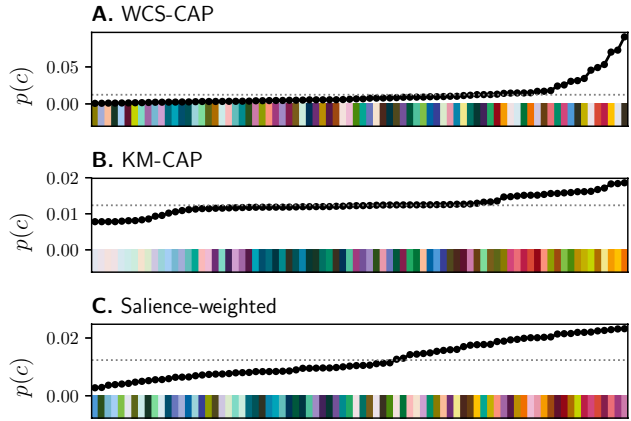

Figure 4: Inferred (**A**: WCS-CAP, **B**: KM-CAP) and directly measured (**C**: salience-weighted) priors. Chips along the $x$-axis are rank ordered according to $p(c)$. Dashed line corresponds to a uniform prior. KM-CAP and salience-weighted exhibit a warm-cool asymmetry, whereas WCS-CAP exhibits a weaker tendency for warm colors and the two most needed colors according to this prior correspond to light and dark.

We considered three different priors, and assessed their effects in analyses of a single dataset, WCS+. We inferred a capacity-achieving prior from the WCS+ data itself (WCS-CAP, Figure 4A): this is an idealized prior that is implicit in these actual color naming systems. We similarly inferred a capacity-achieving prior from the artificial naming data explored above that are derived from k-means clustering (KM-CAP, Figure 4B): this is an idealized prior implicit in these artificial systems that are based on perceptual structure alone. In each case, following Zaslavsky et al. (2018), we evaluated the CAP $p_l(c)$ for each language $l$ (real or artificial) with respect to its color naming distribution $p_l(w|c)$, and averaged together these language-specific priors in order to infer a universal need distribution.[3] That is, we defined

$$p(c) = \frac{1}{L} \sum_{l=1}^{L} p_l(c), \qquad (7)$$

where $L = 111$ is the number of languages in the WCS+ dataset.

For comparison with these inferred priors, we also considered the salience-weighted prior of Gibson et al. (Figure 4C), which is not inferred but is instead grounded directly in the frequency with which colors appear in foreground objects vs. backgrounds in natural images. For each of these three priors — WCS-CAP, KM-CAP, and salience-weighted — we entered it as $p(c)$ into equation (2), and then used equation (1) to obtain the expected surprisal $S(c)$ for each language in the WCS+ data given that prior. We then assessed each prior in two ways: first by asking whether we obtain the CAP-predicted linear relationship between $-\log p(c)$ and $S(c)$, and second by sorting chips by $S(c)$ and asking whether we observe the warm-cool surprisal asymmetry reported by Gibson et al. and also seen in our Figure 3.

[3]We leave for later investigation the interesting question of language-specific need influences.

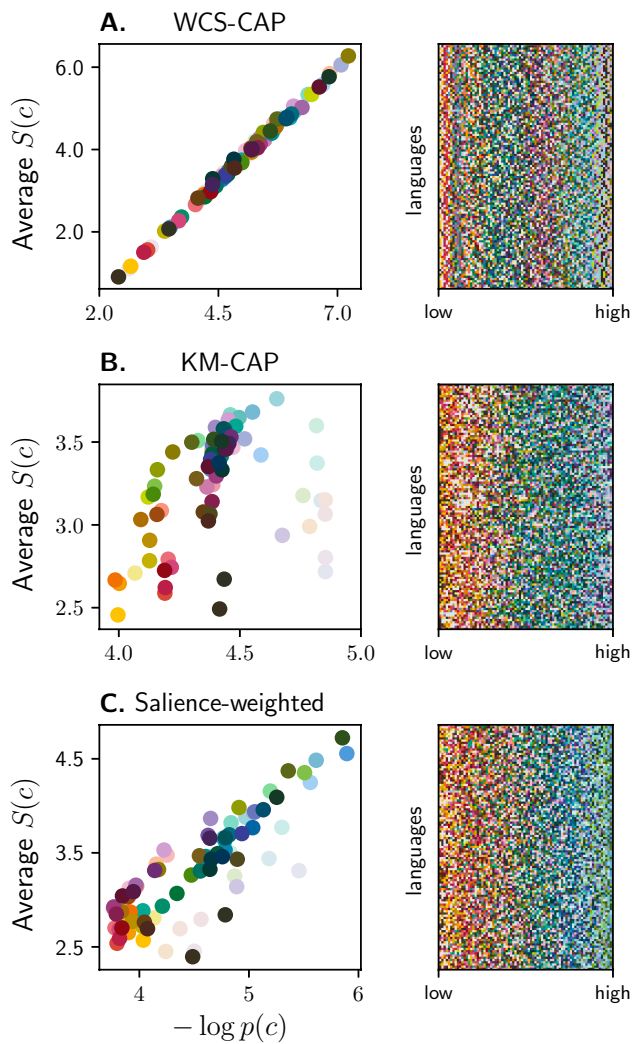

Figure 5: Comparison between the priors of Figure 4. **Left panels:** Scatterplots of $-\log p(c)$ vs. average $S(c)$ across languages. **Right panels:** Surprisal patterns for each prior, analogous to Figure 3. See text for interpretation.

The results are shown in Figure 5. Comparing first just the two inferred priors, WCS-CAP and KM-CAP, we see that the linear relation between $-\log p(c)$ and average $S(c)$ is dissociable from the warm-cool surprisal asymmetry: WCS-CAP shows a linear relation but not a clear warm-cool asymmetry, whereas KM-CAP shows a clear warm-cool asymmetry but not a clear linear relation ($r = 0.32$, $p < 0.01$). The presence of a very clean linear relation for WCS-CAP reassures us that by averaging the language-specific CAPs, we inferred a universal need distribution largely consistent with equation (6).[4] It is perhaps more surprising that the warm-cool asymmetry vanishes under this well-motivated prior, given that it has persisted under others (recall Figure 3). The absence of the warm-cool surprisal asymmetry under WCS-CAP demonstrates the sensitivity of this asymmetry to the assumed prior. At the same time, the lack of a clear linear relation between $-\log p(c)$ and average $S(c)$ under KM-CAP suggests that this prior is not well-suited for precise communication using the naturally occurring color naming systems of the WCS+ dataset. KM-CAP is ultimately derived from perceptual structure, whereas WCS-CAP is derived from the actual WCS+ languages, and both priors are derived using the same principle. Thus, the difference between them, seen in Figures 4 and 5, can be attributed to features in the WCS+ data that are not simply a reflection of perceptual structure.

With this by way of stage-setting, consider now the results for the salience-weighted prior. It exhibits a warm-cool surprisal asymmetry on the WCS+ data (in fact, this panel simply replicates Figure 3B), and also exhibits a roughly linear relation between $-\log p(c)$ and average $S(c)$, with slope close to 1 ($r = 0.83$, $p < 0.0001$). This linear relation is significant for two reasons. First, the fact that this relation is found for the salience-weighted prior but not for the perceptually-based KM-CAP suggests that the salience-weighted prior (like WCS-CAP) exhibits signs of need beyond what is predicted by perceptual structure. Second, this roughly linear relation demonstrates an information-theoretic fit between cross-language color naming data and this prior, which was independently empirically obtained by Gibson et al.

## Discussion

As stated in their title, Gibson et al. (2017) argued that "color naming across languages reflects color use." They presented this claim as an alternative to accounts of color naming based on perceptual salience. In support of this claim, they presented evidence of a warm-cool asymmetry in communicative need and a corresponding asymmetry in communicative precision in color naming across languages — suggesting that color naming systems may have adapted to a universal human tendency to communicate preferentially about warm colors. Here, we have cast this argument in a new light. We have shown that both qualitative asymmetries may be alternatively explained by a common cause: the structure of perceptual color space. Therefore, these two asymmetries are not an unambiguous sign that color naming reflects communicative need.

However, by invoking an information-theoretic principle that links need and precision, we have also presented a dif-ferent form of evidence that color naming does in fact bear traces of universal patterns of communicative need beyond what perceptual factors would predict. Thus, we agree with Gibson et al. that communicative preferences appear to have left their imprint on color naming systems in the world's languages (see also Kemp & Regier, 2012 for a similar argument concerning kin terminologies). However, we differ with Gibson et al. in two respects: first, we reach this conclusion on different grounds, and second, we find that communicative need may operate in concert with, rather than as an alternative to, perceptual structure as a determinant of color naming.

---
[4]By substituting WCS-CAP into equation (6) we introduced a non-linearity because the language-specific CAPs are averaged inside the log. In principle, this could have violated equation (6).

More broadly, there is also another possible connection between perceptual structure and need. Although we have treated these two as independent factors, it may be the case that the structure of perceptual color space is itself adapted to the statistics of natural scenes (Shepard, 1994), and is in that sense influenced by need. Even in this case, however, the picture is not entirely straightforward. There is an important distinction in principle, and thus at least possibly in practice, between the frequency with which particular colors appear in the world, and the frequency with which they must be communicated. It seems likely that our perceptual systems may have adapted to the former, and our languages to the latter.

## Acknowledgments

We thank Bevil Conway and Ted Gibson for kindly sharing their salience-weighted prior with us, Delwin Lindsey and Angela Brown for kindly sharing their English color naming data with us, and Joshua Abbott for helpful discussions. This study was supported by the Gatsby Charitable Foundation (N.Z. and N.T.) and DTRA award HDTRA11710042 (T.R.). Part of this work was done while N.Z. and N.T. were visiting the Simons Institute for the Theory of Computing.